\documentclass[10pt, a4paper]{article}

\usepackage[final]{lrec2026} %

\usepackage{graphicx}
\usepackage{float}
\usepackage{stfloats}
\usepackage{tcolorbox}
\usepackage{xcolor}
\usepackage{listings}
\tcbuselibrary{listings,breakable,skins}
\usepackage{amsmath,amssymb}
\usepackage{todonotes}
\usepackage{url}
\usepackage{booktabs}

\title{Grounded Misunderstandings in Asymmetric Dialogue: A Perspectivist Annotation Scheme for MapTask}

\name{Nan Li\textsuperscript{1}, Albert Gatt\textsuperscript{1}, Massimo Poesio\textsuperscript{1,2}} 

\address{\textsuperscript{1}Utrecht University, Utrecht, the Netherlands\\
        \textsuperscript{2}Queen Mary University of London, London, the UK\\
         \{n.li, a.gatt, m.poesio\}@uu.nl\\}

\abstract{
Collaborative dialogue relies on participants incrementally establishing common ground, yet in asymmetric settings they may believe they agree while referring to different entities.
We introduce a perspectivist annotation scheme for the HCRC MapTask corpus (Anderson et al., 1991) that separately captures speaker and addressee grounded interpretations for each reference expression, enabling us to trace how understanding emerges, diverges, and repairs over time.
Using a scheme-constrained LLM annotation pipeline, we obtain 13k annotated reference expressions with reliability estimates and analyze the resulting understanding states.
The results show that full misunderstandings are rare once lexical variants are unified, but multiplicity discrepancies systematically induce divergences, revealing how apparent grounding can mask referential misalignment.
Our framework provides both a resource and an analytic lens for studying grounded misunderstanding and for evaluating (V)LLMs’ capacity to model perspective-dependent grounding in collaborative dialogue.

 \\ \newline  }

\begin{document}

\maketitleabstract

\section{Introduction}\label{sec:intro}

In everyday collaborative dialogue, speakers and addressees continuously coordinate their understanding of what is being discussed \cite{clark1991GroundingCommunication, pickering2004MechanisticPsychologyDialogue}. A central mechanism for this coordination involves reference expressions (REs)—linguistic forms used to pick out entities in the shared context. 

Utterances become part of the common ground only after being recognized and acknowledged by the addressee~\cite{clark1986ReferringCollaborativeProcess,clark1989ContributingDiscourse,clark1991GroundingCommunication}, a process known as \emph{grounding}. Traditional approaches to reference resolution assume that once grounding is achieved via explicit or tacit confirmation, both interlocutors successfully refer to the same entity. However, this assumption can fail in \emph{asymmetric} settings where participants have access to different information.

\begin{figure}
    \centering
    \includegraphics[width=\linewidth]{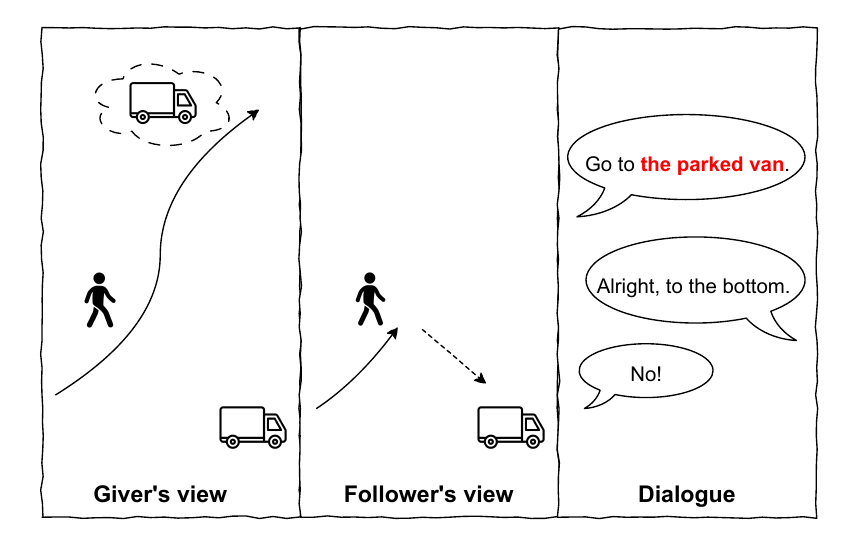}
    \vspace{-5mm}
    \caption{A simplified example of misunderstandings from MapTask dialogues. \textbf{Left}: the instruction \emph{giver}'s map contains two instances of \emph{parked van} and the intended target is circled. \textbf{Middle}: the instruction \emph{follower}'s map shows only one \emph{parked van} at the bottom that is shared with the \emph{giver}. \textbf{Right}: the \emph{giver} says ``go to \emph{the parked van},'' the \emph{follower} acknowledges but grounds another instance, leading to correction (``No!''). Red highlights the reference expression.}
    \label{fig:maptask-example}
\end{figure}

Such asymmetric settings were the focus of the
HCRC MapTask (~\citealp{anderson1991hcrc}; see Figure~\ref{fig:maptask-example} for an example), where two participants navigate a route using slightly different maps. While participants in such settings show systematic adaptation behaviors to address information asymmetry~\cite{bard2000ControllingIntelligibilityReferringa, viethen2011generating, viethen2011ImpactVisualContext, healey2018RunningRepairsCoordinating}, making full misunderstandings rare, few studies have examined the \emph{personal interpretations} of REs by the two interlocutors and the subtle divergences that may persist even after apparent grounding.
These limitations also exist in the MapTask annotations (discussed in Section~\ref{sec:scheme-approach}), making them difficult to be used to identify misunderstandings.

To understand how interlocutors achieve and sometimes fail to achieve mutual understanding, we must first capture each participant’s personal interpretation. Taking both perspectives allows us to quantify where apparent alignment hides subtle misalignments and to measure how frequent such misunderstandings actually are under different landmark discrepancy settings.

Prior frameworks for modeling dialogue states offer rich tools to trace how shared understanding is established~\cite{poesio1997ConversationalActionsDiscourse, cooper1999coding,matheson2000ModellingGroundingDiscourse, lascarides2009agreement, schlangen2011general, ginzburg2012InteractiveStance}. However, these symbolic, frame-based representations have not been widely integrated into reference-level annotation schemes, partly because they are difficult for large language models (LLMs) to reproduce in a constrained, schema-driven manner.

We introduce an annotation framework based on the MapTask annotations that explicitly captures the divergence between the speaker's intended and the addressee's interpreted referents during collaborative dialogue.
Our scheme uses a hierarchy of attributes to represent personal interpretations and incremental grounding, making it both linguistically principled and operational for LLM annotation.

Applying this framework to MapTask dialogues with GPT-5~\cite{OpenAI_GPT-5_2025}, we scale annotation across the entire MapTask corpus. While full misunderstandings remain uncommon, our analysis reveals systematic patterns of understanding state transitions. These findings not only shed light on how diverging interpretations emerge and resolve in collaborative tasks but also establish a benchmark for evaluating language-only and vision-language models’ grounding abilities.

Our research questions are:

\emph{\textbf{RQ1} Is it possible to develop a personal-interpretation annotation scheme to capture nuanced understanding states for REs in collaborative tasks like MapTask?}

\emph{\textbf{RQ2} Given conflicting personal interpretations, can we trace how understanding evolves across turns until the successful grounding?}

\emph{\textbf{RQ3} Can LLMs, under a schema-constrained protocol, reliably annotate personal interpretations; and how does their output inform future evaluation of (V)LLMs on grounded dialogue?}

Our contributions include:

\begin{itemize}
    \item We operationalize reference expression grounding as a graded, perspective-dependent alignment process, and propose an annotation scheme at the RE level.
    \item We present a scheme-constrained LLM-in-the-loop annotation pipeline of MapTask dialogues, producing 13k annotated REs with reliability estimates.
    \item We release analysis protocols that associate RE-level annotations with dual-perspective interpretations and incremental understanding states, enabling quantitative study of misalignment dynamics in MapTask dialogues and informing future evaluation of LLMs and VLMs on common grounding.
\end{itemize}

The remainder of this paper is structured as follows. Section~\ref{sec:related_work} reviews related work. Section~\ref{sec:maptask} introduces the MapTask corpus and its challenges for reference grounding. Section~\ref{sec:scheme} details our annotation scheme and LLM-in-the-loop annotation pipeline. Section~\ref{sec:discussion} analyzes understanding state distributions and transitions. Finally, Section~\ref{sec:conclusion} concludes with implications and future directions.

\section{Related work}\label{sec:related_work}

\paragraph{Collaborative Reference and Grounding}

Reference expressions are central to establishing and maintaining common ground in interactive dialogue~\cite{clark1986ReferringCollaborativeProcess, clark1989ContributingDiscourse}. Successful mutual understanding is constrained by many factors, e.g., communication medium, interactive alignment at different linguistic levels~\cite{clark1991GroundingCommunication,pickering2004MechanisticPsychologyDialogue}. Previous work reveals that speakers do not always fully accommodate addressee knowledge, exhibiting egocentric biases even when explicitly aware of information asymmetry~\cite{horton1996speakers, keysar2000taking, lane2006DontTalkPink}
 
This theoretical foundation has motivated NLP tasks to study grounded reference in collaborative settings with asymmetric information. The HCRC MapTask~\cite{anderson1991hcrc} and iMap corpus~\cite{louwerse2007multimodal} both instantiate navigation tasks where participants have slightly different maps, eliciting rich adaptation behavior~\cite{bard2000ControllingIntelligibilityReferringa, viethen2011generating, viethen2011ImpactVisualContext}. More recent work includes the OneCommon Corpus~\cite{udagawa2019NaturalLanguageCorpus}, a partially-observable environment for both participants. These approaches typically treat grounding as a single end state rather than tracking divergent personal interpretations throughout the dialogue.

Current VLMs have excelled at some object identification tasks \cite{yu2016modeling}, where referents (objects in images) are assumed to be deterministically grounded once specified. Some recent work shows that VLMs struggle to adopt different frames of reference and visual perspectives~\cite{zhang2024vision, lee2025perspective}. Moreover, existing benchmarks rarely evaluate whether models can track evolving, asymmetric interpretations as dialogue progresses. Our work addresses this gap by annotating personal interpretations at the RE level, enabling future evaluation of whether (V)LLMs understand incremental grounding processes in (asymmetric) dialogue.

\paragraph{Understanding State Tracking} 

Meaning coordination operates through continuous ``running repairs'' rather than discrete error correction episodes~\cite{healey2018RunningRepairsCoordinating}. This repair-driven coordination directs the participants' evolving understanding states, and motivates the need to track divergent interpretations as they emerge and resolve through integration.

Formal dialogue theories have modeled interlocutors' intentions and repairs through structured information state models. 
Some proposals~\cite{poesio1997ConversationalActionsDiscourse, matheson2000ModellingGroundingDiscourse} extended  Discourse Representation Theory to integrate semantic and pragmatic information, showing how turn-taking, discourse segmentation, and grounding are represented through conversation acts in common ground.
\citet{ginzburg2012InteractiveStance} proposed the KoS framework\footnote{A term loosely connected to \emph{conversation-oriented semantics}~\cite{ginzburg2012InteractiveStance}.}, where unresolved Questions Under Discussion (QUDs) persist until explicitly addressed.
\citet{lascarides2009agreement} modeled agreement and correction dynamics but assumed a highly idealized scenario which excludes misunderstandings.
\citet{schlangen2011general} treated dialogue processing as an incremental hypothesis-formation and revision process operating at a sub-utterance level.
Some recent work leveraged annotations from other meaning represenatations, e.g., Abstract Meaning Representation, to represent common ground structure and information states~\cite{khebour2024CommonGroundTracking,lai2025model}. We are informed by this line of work, but focus on RE-level perspectivist interpretations and inferring understanding states through the alignment and misalignment of these interpretations.

\paragraph{LLM as an Annotator} LLMs have demonstrated strong performance on various semantic and pragmatic annotation tasks~\cite{eichin2025probing, chen2025semantic, qamar2025llms}.
They also achieve comparable results to human annotators in subjective tasks such as sentiment analysis~\cite{bojic2025comparing}.
However, challenges remain for scheme-constrained annotation tasks requiring structured outputs~\cite{ettinger-etal-2023-expert}.
Recent advances in grammar-constrained decoding and JSON schema-based generation offer promising solutions, enabling LLMs to generate outputs that adhere to predefined structures while maintaining semantic accuracy~\cite{geng-etal-2023-grammar, park2025flexible}.
Our work builds on these developments by employing a multi-layer attribute resolution pipeline to guide LLMs in capturing personal interpretations in collaborative dialogue.

\section{MapTask: Annotation and Beyond}\label{sec:maptask}

\subsection{Development and Follow-up Studies}

HCRC MapTask is a corpus documenting unscripted dialogues in a collaborative route-following task with asymmetric maps. Participants must collaborate to reproduce a route on one participant's map (the instruction \emph{giver}'s) on the other's (the instruction \emph{follower}'s). They cannot see each other's maps and can only communicate verbally, with eye contact possible in half of the dialogues. The intentional differences between maps on routes, landmark names, and landmark placements often lead to ambiguous or misleading instructions. This makes the MapTask corpus a valuable resource and testbed for studying grounding, reference, and repair theories in dialogue.

Subsequent research analyzed reference expressions in MapTask from multiple perspectives.
\citet{bard2000ControllingIntelligibilityReferringa} show that speakers reduce articulatory clarity when mentioning landmarks for the second time, even when the addressee has not heard the first mention or cannot see the referent, suggesting that intelligibility control is driven primarily by the speaker's own knowledge rather than careful modeling of the listener's information state.
\citet{varges2005spatial} treated spatial descriptions in MapTask as reference expressions that distinguish particular map points from distractors, observing that speakers frequently use spatial relations (e.g., ``above the west lake'') and vague quantifiers (e.g., ``about two inches southwest'') when referring to featureless route points around named landmarks.
\citet{marge2011towards} analyzed miscommunication patterns in MapTask, documenting how instruction followers detect ambiguity (e.g., multiple possible referents) and misunderstanding through clarification questions, repair strategies, and explicit feedback when requested actions appear impossible or unclear.
\citet{Poesio2004TheVC} applied the MATE annotation scheme \cite{poesio1999mate} to an Italian adaptation of MapTask to make anaphoric relations explicit.

\subsection{Task Design and Landmark Discrepancies}\label{sec:maptask-design}

The HCRC MapTask corpus consists of 16 pairs of maps and 128 dialogues, with 8 dialogues recorded for each map. In each dialogue, two participants are given visually similar but non-identical maps. One participant, the \emph{giver}, has a route marked on the giver's map and must guide the other participant, the \emph{follower}, to reproduce this route on the follower's map through verbal communication alone. Both maps share the same starting point and most landmarks, but only the giver's map displays the complete route and endpoint. Critically, participants cannot see each other's maps, creating an asymmetric information setting.

The corpus contains 267 unique landmarks (differentiated by names) across all maps. While both participants' maps share many identical landmarks (132 landmarks, 49.4\%), the remaining 135 landmarks (50.6\%) exhibit systematic discrepancies. We classify these discrepancies into three categories~\footnote{In the original MapTask paper~\cite{anderson1991hcrc}, the authors claim that they also add some phonetic contrast features to landmark names to create discrepancies; we assign them to the following three categories according to our definition and observation.} and present the distribution by discrepancy type in Table~\ref{tab:landmark_dist}.

\begin{table}[htb]
\centering
\resizebox{\linewidth}{!}{
\begin{tabular}{lrr|rr}
\toprule
\textbf{Discrepancy Type} & \textbf{Count} & \textbf{\%} & \textbf{Ref.} & \textbf{\%}\\
\midrule
Lexical & 20 & 7.5  & 914 & 7.0 \\
Existence & 99 & 37.1  & 2878 & 22.0\\
Multiplicity & 16 & 6.0  & 956 & 7.3\\
\midrule
Discrepant & 135 & 50.6  & 4748 & 36.3\\
Identical & 132 & 49.4  & 8333 & 63.7\\
\midrule
Total & 267 & 100.0 & 13081 & 100.0\\
\bottomrule
\end{tabular}}
\caption{Distribution of landmark types and corresponding reference expressions in the MapTask corpus.}
\label{tab:landmark_dist}
\end{table}

\paragraph{Lexical Discrepancy} (20 landmarks forming 10 pairs, 7.5\%): Landmarks appear at the same location on both maps with identical icons but slightly different name labels. For instance, one map may label a landmark ``cliffs'' while the other reads ``sandstone cliffs''; or ``carved wooden pole'' versus ``totem pole''. 

\paragraph{Existence Discrepancy} (100 landmarks, 37.5\%): Landmarks appear on one participant's map but are entirely absent from the other's. The participant viewing the map without the landmark must either signal the absence or imagine its location based on the partner's description.

\paragraph{Multiplicity Discrepancy} (16 landmarks, 6.0\%): Landmarks appear twice on one map (always the giver's) but only once on the other (the follower's). Shared instances of this type occupy the same location on both maps but are positioned farther from the route than the unique, non-shared instances.

\subsection{Annotations Leveraged in This Work}

The dataset of the MapTask corpus is publicly available online\footnote{\url{https://groups.inf.ed.ac.uk/maptask/}} with rich multi-level annotations. For our experiment and analysis, we leverage the following annotation layers: (1) \textbf{Timed Units}, word-level transcriptions aligned with audio timestamps, providing the basic segmentation of speech and text. (2) \textbf{Reference Expressions}, annotations of reference expressions linked to landmark IDs, recorded based on timed units. (3) \textbf{Move Tags and Transactions}, move-level annotations following the coding scheme of \citet{carletta1997reliability}, capturing the communicative function of utterances (e.g., instruct, acknowledge, query), recorded based on timed units. Transactions are sequences of moves that move from one point to another on maps.

We conducted preprocessing to integrate these annotations into a unified format suitable for LLM annotation. First, we reassigned landmark identifiers (IDs) to distinguish between giver-side and follower-side interpretations (in Section~\ref{sec:scheme-approach}), as the original landmark ID scheme fails to highlight the discrepancies in some cases. Second, we segmented the dialogue transcripts into transactions to control the context window when providing context to the model. Finally, we generated structured prompts combining reference expressions, new unified landmark ID candidates, and context dialogue to support the annotation workflow described in Section~\ref{sec:scheme-pipeline}.

\section{Modeling Personal Interpretations}\label{sec:scheme}

\subsection{Approach and General Principles}\label{sec:scheme-approach}

We introduce a perspectivist, RE-level annotation scheme that records the speaker-intended and addressee-interpreted landmark IDs, plus five binary attributes in a fixed decision order.

\paragraph{Landmark ID Design}

The original MapTask annotations assign landmark IDs by landmark names, which treat landmarks of multiplicity discrepancies (e.g., two ``parked van'' at the giver's map, one at the follower's map) as the same entity, but treat lexical variants (e.g., ``cliffs'' versus ``sandstone cliffs'') as different entities. This conflation obscures whether participants successfully align their interpretations. To address this limitation, we introduce a new unified landmark ID assigning strategy that distinguishes giver-side and follower-side landmark instances. In this paper, we use the acronyms \verb|mtlm| and \verb|umlm| to refer to the original MapTask landmark ID and our unified landmark ID, respectively.

Our landmark IDs follow the format \texttt{\small <map-id>\_<landmark-name>\#<ordinal>@<side>}, where:
\begin{itemize}
\item \texttt{map-id}: The original Map ID (\texttt{m0}, \texttt{m1}, \ldots, \texttt{m15})
\item \texttt{landmark-name}: The landmark name as it appears on the map (e.g., \texttt{parked\_van}, \texttt{cliff})
\item \texttt{ordinal}: Positional index (\texttt{0}, \texttt{1}, \ldots) for landmarks of the same type occurring multiple times on maps, ordered from bottom to top. Landmarks of this type have the same ordinal number if they are at the same location on both maps, which means it is possible that the only ``parked van'' on the follower's map is \texttt{m0\_parked\_van\#1@f} if the giver's map has two ``parked van'', while on the giver's map \texttt{m0\_parked\_van\#0@g} denotes the unique instance that is at a lower place.
\item \texttt{side}: Map side (\texttt{g} for giver, \texttt{f} for follower)
\end{itemize}

When a reference expression refers to multiple landmarks simultaneously, we concatenate IDs with the operator \texttt{+}, e.g., \texttt{m0\_parked\_van\#0@g+m0\_parked\_van\#1@g}.

Beyond landmark IDs, we annotate five binary attributes that jointly characterize the nature of the expression and the addressee's understanding state. These attributes are designed in a hierarchical decision cascade that models the incremental resolution of reference and embeds Chain-of-Thought \cite{wei2022chain} into LLMs' reasoning process.

\paragraph{\texttt{is\_quantificational}} Indicates whether the reference expression functions as an existence query from the speaker's perspective rather than a definite reference. For instance, ``a parked van'' in ``Do you have a parked van?'' is quantificational, whereas ``the parked van'' in ``Go past the parked van'' is not. Such quantificational expressions do not typically presuppose a specific referent in MapTask contexts.

\paragraph{\texttt{is\_specified}} Indicates whether the dialogue context provides sufficient evidence to infer the addressee's interpretation. If the addressee fails to respond or produces an utterance unrelated to the reference expression, the interpretation remains unspecified.

\paragraph{\texttt{is\_accommodated}} Indicates whether the addressee successfully address the reference expression without signaling comprehension failure. Accommodation fails when the addressee explicitly requests repetition, clarification, or indicates mishearing (e.g., ``What?'' or ``Pardon?'').

\paragraph{\texttt{is\_grounded}} Indicates whether the addressee links the reference expression to a specific landmark on the maps. An accommodated expression may fail to ground if the addressee cannot identify a corresponding landmark (e.g., ``I don't have that'' or ``Where is it?''). Grounding is a prerequisite for assigning a landmark ID to the addressee's interpretation.

\paragraph{\texttt{is\_imagined}} Indicates whether the addressee's interpretation refers to a landmark absent from his/her own map. This occurs when the addressee, having recognized an existence discrepancy, mentally projects the landmark's location based on the speaker's description. In such cases, we assign the speaker's landmark ID to the addressee's interpretation and mark \texttt{is\_imagined = true}, signaling that alignment was achieved.

\subsection{Annotation Pipeline}\label{sec:scheme-pipeline}

\begin{figure*}[htb]
\centering
\begin{tcolorbox}[
  title=Prompt Template (Abbreviated),
  colback=gray!5,
  colframe=black!75,
  fonttitle=\bfseries\small,
  left=2mm,
  right=2mm,
  top=1mm,
  bottom=1mm
]
{\scriptsize
\texttt{<background>} Background on MapTask, landmark discrepancies, and reference resolution \texttt{</background>}\\[2pt]
\texttt{<task\_description>} Annotate personal interpretations for reference expressions \texttt{</task\_description>}\\[2pt]
\texttt{<landmark\_id\_explanation>} Unified landmark ID format and semantics \texttt{</landmark\_id\_explanation>}\\[2pt]
\texttt{<annotation\_rule>}\\
\quad Follow the annotation workflow for each RE step by step:\\
\quad Step 0: Identify the speaker and the addressee, and match them with the giver and follower based on the actual situation; fill in the landmark ID for the speaker's interpretation (intended landmark) first.\\
\quad Steps 1-5: Check and fill in attributes is\_quantificational, is\_specified, is\_accommodated, is\_grounded, and is\_imagined in order.\\
\quad Step 6: If grounded, fill the landmark ID for the addressee's interpretation.\\
\quad Step 7: Provide a concise (<50 words) evidence-based reason summarizing the judgement in the reason field for each RE.\\
\texttt{</annotation\_rule>}\\[2pt]
\texttt{<output\_format>} JSON schema with required fields \texttt{</output\_format>}\\
\texttt{<target\_ref\_ids>} IDs of reference expressions to annotate \texttt{</target\_ref\_ids>}\\
\texttt{<context\_dialogue>} Dialogue transcript with surrounding context \texttt{</context\_dialogue>}\\
\texttt{<context\_dialogue\_acts>} Move-level annotations for context \texttt{</context\_dialogue\_acts>}\\
\texttt{<landmark\_candidates>} All landmarks on both maps with IDs \texttt{</landmark\_candidates>}
}
\end{tcolorbox}
\caption{Prompt template for annotation, abbreviated due to space limits. Figure~\ref{fig:app-prompt} in Appendix~\ref{sec:app-prompt} provides a more detailed version of the prompt template.}
\label{fig:prompt}
\end{figure*}

Informed by recent advances in Prompt Engineering \cite{brown2020language,wei2022chain,sahoo2024systematic}, we develop a heuristic prompting annotation pipeline leveraging LLMs to assist with the perspective-taking annotation task.
To seek a balance between information density and context length \cite{liu2023lost,kuratov2024babilong}, we provide the dialogue text from the start of the conversation until the end of the current transaction as context.
We use a scheme-constrained prompt with GPT-5 via the OpenAI Batch API\footnote{\url{https://platform.openai.com/docs/guides/batch}} and enforce a JSON-schema output to scale annotation with quality control. All parameters are set to default.

For each reference expression, we construct a structured prompt containing:

\begin{itemize}
\item \textbf{Background} Overview of the MapTask task design and landmark discrepancy types, and a general introduction of the annotation task.
\item \textbf{Task description}: Detailed description of the annotation task, emphasizing the approach of using landmark IDs and 5-layer attribute hierarchy to annotate.
\item \textbf{Landmark ID explanation}: Explanation of the unified landmark ID format.
\item \textbf{Annotation rule}: Step-by-step workflow instructions operationalizing the five-attribute decision and landmark ID resolution cascade.
\item \textbf{Output format}: JSON schema with required fields.
\item \textbf{Target reference expressions}: All reference expressions in the current transaction to annotate with their IDs.
\item \textbf{Dialogue context}: Dialogue transcript from the start of the conversation until the end of the current transaction, with reference expressions highlighted by brackets.
\item \textbf{Dialogue acts}: Move-level annotations indicating communicative function in time order, e.g. "[g utt:1 move:instruct]     starting off we are above a caravan park [f utt:2 move:acknowledge]  mmhmm".
\item \textbf{Landmark candidates}: List of landmarks on maps with their \verb|umlm| IDs, based on the scheme defined in the previous section.
\end{itemize}

The prompt (see Figure~\ref{fig:prompt} and \ref{fig:app-prompt}) guides the model through a sequential and heuristic annotation workflow. First, the model identifies the speaker and addressee roles (giver/follower) and assigns the speaker's intended landmark ID. Second, it evaluates \texttt{is\_quantificational} to determine whether the expression presupposes a referent. If not quantificational, the model proceeds through the remaining four attributes in order. Finally, if \texttt{is\_grounded=true}, the model assigns the addressee's interpreted landmark ID. The model is asked to give a concise textual reason citing dialogue evidence for our further analysis.

\subsection{Quality Evaluation}\label{sec:scheme-eval}

We apply the annotation pipeline to all 128 dialogues in the MapTask corpus, resulting in 13,077 annotated reference expression instances~\footnote{The LLM omitted annotations for 4 REs (from 3 different dialogues), despite their presence in both the dialogue context and the RE list, likely due to uncertainty during the generation process.}.

To evaluate the reliability of the annotation pipeline, we selected 3 dialogues (dialogue IDs: \verb|q1ec2|, \verb|q1nc3|, \verb|q1nc7|) for complete human verification, comprising 504 reference expressions. We compute micro-averaged accuracy and f1-scores for each binary attribute by comparing LLM predictions against the human gold standard. The human gold standard was collected by one experienced annotator (one of the paper authors) following the same annotation guidelines, with ambiguous cases discussed and resolved with other authors.

\begin{table}[htbp]
\centering
\resizebox{\linewidth}{!}{
\begin{tabular}{lrrrr}
\toprule
\textbf{Attribute} & \textbf{N} & \textbf{Acc.} & \textbf{F1} & \textbf{Errors}\\
\midrule
is\_quantificational & 504 & 100.00\% & 1.00 & 0\\
is\_specified & 466 & 97.85\% & 0.990 & 10\\
is\_accommodated & 455 & 97.58\% & 0.988 & 11\\
is\_grounded & 447 & 96.20\% & 0.980 & 17\\
is\_imagined & 422 & 97.63\% & 0.891 & 10\\
\midrule
\multicolumn{4}{l}{\textbf{RE-level}: 28/504 REs (5.6\%) with $\geq$1 error} & 48\\
\bottomrule
\end{tabular}}
\caption{Evaluation metrics by attribute across 3 dialogues (q1ec2, q1nc3, q1nc7; micro-averaged). N indicates the number of REs evaluated at each attribute (determined by gold annotations). Bottom line reports the RE-level error count (unique REs with $\geq$1 attribute errors), which reflects an accuracy of 94.4\%.}
\label{tab:human_eval_metrics}
\end{table}

The LLM annotations exhibit relatively high validity across multiple evaluation dimensions when compared to human gold standard.
For all grounded reference expressions, the LLM achieves 95.5\% accuracy and 99.5\% micro-F1, demonstrating reliable recovery of each participant's intended and interpreted referents.
Table~\ref{tab:human_eval_metrics} then presents the micro-averaged evaluation results for the five binary attributes across all 3 dialogues.
The LLM achieves high performance across all attributes, with accuracy ranging from around 97\% to 100\% and F1 scores from 0.89 to 1.00.
At the reference expression level, 28 of 504 REs (5.6\%) contained at least one attribute error, totaling 48 attribute-level mismatches. 
Attribute-level error breakdown shows that \texttt{is\_grounded} has the highest error count (17), largely stemming from \emph{false negatives} where the RE is grounded in gold but missed by the model, which typically occur when grounding is inferred from implicit confirmation or reuse rather than explicit echoing.

\section{Discussion}\label{sec:discussion}

\subsection{Classification of Understanding States}\label{sec:scheme-typology}

Our annotation framework derives understanding states from hierarchical attribute decisions at the RE level. We define three primary categories that capture the alignment or divergence between speaker and addressee interpretations:

\paragraph{Aligned} The addressee successfully accommodates and grounds the reference expression to the speaker's intended landmark, and the landmark \verb|mtlm| IDs match between speaker and addressee interpretations. This represents successful reference resolution.

\paragraph{Misunderstood} The addressee grounds the reference expression to a different landmark from that the speaker intended. Both participants believe successful reference has occurred, yet their interpretations diverge. Such silent misalignments may persist undetected unless subsequent dialogue reveals the discrepancy.

\paragraph{Pending} The interlocutors await further information or signal comprehension failure. This category includes several communication breakdown scenarios corresponding to the attributes discussed in Section~\ref{sec:scheme-approach}, except for \verb|is_imagined|.

For each RE, we infer the interlocutors' understanding state from our annotated data, and Table~\ref{tab:understanding_dist} presents the distribution.
The ``misunderstood'' type reaches 7.07\%. However, as discussed in Section~\ref{sec:maptask-design}, 7.5\% of landmarks exhibit \emph {lexical discrepancies} where the same landmarks are labeled with different names on maps. For instance, when the giver says ``the old mill'' (referring to \texttt{m12\_old\_mill@g}) and the follower only has \texttt{m12\_mill\_wheel@f} on his/her map, they may initially correct each other but usually quickly reach an agreement that they are referring to the ``same'' landmark at the same location. In such cases, despite the initial lexical mismatch, we assume they have reached an aligned understanding state. 

Verified by our observations, we unified the 10 pairs of lexical variants and adjust the ``misunderstood'' cases caused by them into ``Aligned'' ones. Table~\ref{tab:understanding_dist} demonstrates the understanding state distribution after this unification as well. Now the misunderstanding rate drops to 1.82\%, a significant reduction from the original 7.07\%.

\begin{table}[h]
\centering
\resizebox{\linewidth}{!}{
\begin{tabular}{lrr|rr}
\toprule
\textbf{State} & \multicolumn{2}{c|}{Before} & \multicolumn{2}{c}{After}\\
& \textbf{Count} & \textbf{\%} & \textbf{Count} & \textbf{\%}\\
\midrule
Aligned & 8,750 & 66.9 & 9,435 & 72.1\\
Pending & 3,403 & 26.0 & 3,403 & 26.0\\
Misunderstood & 924 & 7.07 & 239 & 1.82\\
\midrule
Total REs & 13,077 & 100.0 & 13,077 & 100.0\\
\bottomrule
\end{tabular}}
\caption{Distribution of understanding states before and after lexical variant unification in the LLM-annotated MapTask reference expression dataset.}
\label{tab:understanding_dist}
\end{table}

The relatively low misunderstanding rate aligns with \citet{healey2018RunningRepairsCoordinating}'s analysis of rarity of sustained misunderstandings in collaborative dialogue: participants actively monitor alignment through continuous ``running repairs'' and employ clarification strategies to prevent persistent divergence.

\begin{table}[h]
\centering
\resizebox{\linewidth}{!}{
\begin{tabular}{lrrr}
\toprule
\textbf{Discrepancy Type} & \textbf{Total REs} & \textbf{Misunderstanding} & \textbf{Rate (\%)}\\
\midrule
Lexical & 914 & 13 & 1.4\\
Multiplicity & 956 & 115 & 12.0\\
Existence & 2,878 & 93 & 3.2\\
Identical & 8,333 & 18 & 0.2\\
\midrule
Total & 13,081 & 239 & 1.8\\
\bottomrule
\end{tabular}}
\caption{Misunderstanding counts and rates by landmark discrepancy type.}
\label{tab:misunderstanding_by_type}
\end{table}

Table~\ref{tab:misunderstanding_by_type} demonstrates how these misunderstandings distribute across different types of landmark discrepancies.
The distribution shows a pattern: multiplicity discrepancies account for 48.5\% of all misunderstandings (115/239 cases) despite representing only 7.3\% of reference expressions in the corpus. The misunderstanding rate of this type is more than 6$\times$ higher than the corpus average (12.0\% versus 1.8\%).
This reveals the success of the MapTask asymmetric design in eliciting misunderstandings, as multiplicity discrepancies inherently create ambiguity about which specific landmark instance is being referred to: when a landmark type appears twice on the giver's map but only once on the follower's map, participants often assume referential uniqueness \cite{lane2006DontTalkPink,bard2000ControllingIntelligibilityReferringa}.
For example, the giver refers to instance \#0, the follower accommodates to their visible instance \#1, and both believe successful reference has occurred. This silent divergence persists unless subsequent spatial descriptions reveal the mismatch.

For ``misunderstood'' cases within the existence discrepancy and identical landmarks, we find that the most frequently occurring cause is still lexical similarity. For example, ``white mountain'' on Map \verb|m9| with the highest misunderstanding occurrences in the identical type is usually confused with ``slate mountain'' on the same map, which has the highest misunderstanding occurrences in the existence discrepancy type. ``Rock fall'' with the second highest misunderstanding occurrences in the existence discrepancy type is often misunderstood as ``stones'' on the same map. These landmarks all have different names, icons, and locations.

\subsection{Reference Chain}\label{sec:chain-discussion}

To analyze how understanding evolves across repeated references to the same landmark, we extracted \emph{reference chains}—sequences of reference expressions targeting the same landmark (same \verb|mtlm|) within a dialogue. Across 128 dialogues, we identified 1,665 reference chains with a mean length of 6.89 REs per landmark (range: 1–70).

\begin{table}[h]
\centering
\resizebox{\linewidth}{!}{
\begin{tabular}{lrrrr}
\toprule
\textbf{Discrepancy Type} & \textbf{\# Chains} & \textbf{Mean Length} & \textbf{Max Length} & \textbf{Median Length}\\
\midrule
Identical & 939 & 8.16 & 70 & 6\\
Lexical & 79 & 10.70 & 61 & 9\\
Existence & 539 & 4.09 & 31 & 3\\
Multiplicity & 108 & 7.00 & 26 & 5\\
\bottomrule
\end{tabular}}
\caption{Reference chain length by discrepancy type.}
\label{tab:chain_length_by_type}
\end{table}

The length of these chains partly reflects how persistently participants negotiate understanding of landmarks and use these landmarks to generate spatial descriptions. Table~\ref{tab:chain_length_by_type} summarizes chain lengths by discrepancy type.
Lexical variants yield the longest chains on average (10.7 references), as participants often need multiple turns to resolve naming mismatches and confirm shared understanding.
Landmarks with no discrepancy generate the second longest chains on average, while existence discrepancies yield the shortest.
This suggests that existence discrepancies are usually quickly ruled out as reference anchors when participants realize the landmark is absent from one map, whereas references to non-discrepant landmarks persist longer because they support iterative spatial description and navigation coordination, aligning with previous findings~\cite{viethen2011generating,reitter2014alignment}.

To illustrate how understanding states evolve through a reference chain, we present a simple multiplicity discrepancy case from dialogue \texttt{q7ec2}. The landmark \emph{vast meadow} appears twice on the giver's map (\texttt{m11\_vast\_meadow\#0@g}, lower/unique; \texttt{m11\_vast\_meadow\#1@g}, upper/shared) but only once on the follower's (\texttt{m11\_vast\_meadow\#1@f}, upper/shared).

Table~\ref{tab:chain-example} traces five reference expressions to this landmark (see Figure~\ref{fig:app-chain} in Appendix~\ref{sec:app-chain} for the dialogue excerpt). The giver initiates with ``a meadow'' (\texttt{ref.112}), intending the lower instance (\texttt{\#0@g}). The follower queries ``a vast meadow?'' (\texttt{ref.114}) and the giver confirms (\texttt{ref.115}). However, the follower identifies his/her own, the upper shared, vast meadow (\texttt{\#1@f}), creating a \textbf{misunderstanding}: both have landmarks in mind, yet they are different ones.

\begin{table}[h]
\centering
\resizebox{\linewidth}{!}{
\begin{tabular}{clllll}
\toprule
\textbf{Ref} & \textbf{Spk} & \textbf{Expression} & \textbf{Giver ID} & \textbf{Follower ID} & \textbf{State}\\
\midrule
112 & G & ``a meadow'' & \texttt{\#0@g} & \emph{ungrounded} & Pending\\
114 & F & ``a meadow'' & \texttt{\#0@g} & \emph{ungrounded} & Pending\\
115 & G & ``a vast meadow'' & \texttt{\#0@g} & \texttt{\#1@f} & \textbf{Misunderstood}\\
116 & F & ``a vast meadow'' & \texttt{\#1@g} & \texttt{\#1@f} & Aligned\\
119 & F & ``vast meadow'' & \texttt{\#1@g} & \texttt{\#1@f} & Aligned\\
127 & G & ``a vast meadow'' & \texttt{\#0@g} & \emph{ungrounded} & Pending\\
\bottomrule
\end{tabular}}
\caption{Reference chain for \emph{vast meadow} in dialogue \texttt{q7ec2}. Landmark IDs (interpretations of the giver and the follower) are abbreviated (full prefix: \texttt{m11\_vast\_meadow}).}
\label{tab:chain-example}
\end{table}

In the next turn (\texttt{ref.116}), the follower describes the location as ``just below the blacksmiths and carpenters,'' which corresponds to the upper meadow shared on both maps. The giver accepts this reframing via confirmation, and both participants align on the upper instance (\texttt{\#1}). This alignment persists through \texttt{ref.119}. However, later in the dialogue (\texttt{ref.127}), the giver again mentions ``a vast meadow'' in a different spatial context, referring back to the lower instance (\texttt{\#0@g}), but the follower cannot ground it, reverting the state to \emph{pending}.

This chain exemplifies how multiplicity discrepancies create recurring alignment challenges: the same landmark name maps to different physical locations on the maps for participants, producing misunderstandings that may resolve through spatial negotiation but can resurface later in the dialogue.

\subsection{Turns to Ground}\label{sec:turns}

\begin{figure}[h]
    \centering
    \includegraphics[width=\linewidth]{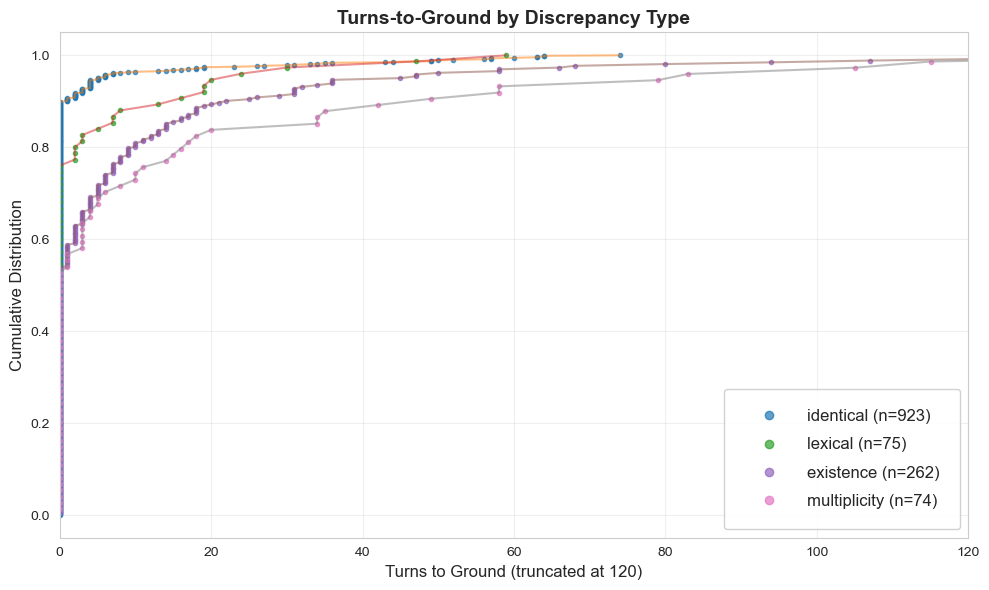}
    \vspace{-5mm}
    \caption{Cumulative distribution of turns-to-ground by discrepancy type.}
    \label{fig:turns_to_ground_plot}
\end{figure}

To further disentangle the efforts participants invest in coordinating understanding of landmarks from using grounded landmarks for navigation, we measure the number of turns (counted per utterance) between a chain's first reference and its first aligned state, and term this measure \emph{turns-to-ground}.
Figure~\ref{fig:turns_to_ground_plot} shows the cumulative distribution of turns-to-ground by discrepancy type across 1,334 chains that reached alignment. The longer negotiation for multiplicity likely reflects the cognitive complexity of distinguishing between multiple same-named landmarks versus simply confirming/denying existence. Participants must coordinate not only on the landmark name but also on which specific instance is intended, requiring more iterative spatial descriptions and confirmations.

There are 331 chains that never reach an aligned state at all, indicating that participants sometimes stop early and abandon negotiation on some landmarks. This aligns with task-oriented dialogue efficiency.

\subsection{Special Case}

In dialogue \texttt{q8ec4}, the giver mistakenly takes the finish as the start point, leading to a persistent spatial misalignment that lasts for nearly half the conversation. The speaker insists on using absolute directions and distances (e.g., ``thirty degrees up the way'' and `` two inches to your left'') to give instructions rather than landmark-relative spatial descriptions, extending the misunderstanding further.

In the annotation, the LLM fails to capture the misalignment since the beginning, partly because of an omniscient bias: the model can ``see'' both maps at once, so it underestimates asymmetric knowledge and treats contradictions as noise rather than evidence of a wrong starting point. Prior studies show that human addressees can achieve more efficient communication than overhearers, while VLMs perform poorly at overhearing~\cite{schober1989UnderstandingAddresseesOverhearers, wang2025LVLMsAreBad}. Future evaluations could explicitly test omniscience and perspective (analyzer versus participant versus overhearer conditions) to test models' perspective-taking ability in grounded dialogue.

\section{Conclusion}\label{sec:conclusion}

Understanding when interlocutors truly ground reference expressions versus only appear to do so requires capturing both sides' interpretations.
This work establishes such a perspectivist view by mapping reference expressions to understanding states in asymmetric collaborative dialogue.
Our annotation scheme disentangles the speaker’s intended and the addressee’s interpreted referents through a five-attribute hierarchy and unified landmark IDs, operationalized via a scheme-constrained prompting pipeline.

Our quantitative analyses reveal that given the rarity of full misunderstandings, multiplicity discrepancies act as a consistent hazard for eliciting misalignment. Our statistics also support prior studies on interlocutors' preferences for reusing grounding-established REs \cite{healey2018RunningRepairsCoordinating, viethen2011generating,reitter2014alignment}.

We plan to establish evaluations based on these annotated dialogues for evaluating (V)LLMs' incremental grounding abilities under different information settings, moving beyond object identification toward perspective-taking in collaborative dialogue.

\section{Limitations}\label{sec:limit}

\paragraph{Annotation Reliability} Our annotation pipeline relies on a single LLM (GPT-5), and the human gold standard was produced by one annotator (with ambiguous cases discussed among co-authors).
This constrains claims about cross-model generalizability and inter-annotator agreement. Although our annotation scheme is desingned as a structured and heuristic decision cascade with JSON-shema validation and explicit explanations --- achieving 94.4\% RE-level accuracy across 504 reference expressions (see Section~\ref{sec:scheme-eval}) --- these results reflect one model under one set of conditions.
Comprehensive evalution across multiple LLMs and human annotators remains an important direction for future work.

\paragraph{Spatial Understanding Precision} Our perspectivist scheme tracks landmark-level interpretations but does not capture finer-grained spatial reasoning. The MapTask involves route navigation between landmarks, yet our annotations focus solely on landmark reference resolution. Path choices, direction ambiguities, and spatial relations between landmarks remain unannotated.

\paragraph{Multimodal information} The original MapTask corpus was collected in conversational settings where participants could perceive prosody, and even have eye contact in half of the cases. Our annotation pipeline relies exclusively on dialogue transcripts and maps, potentially missing non-verbal signals that influence grounding. For instance, backchannel responses like ``mmhmm'' may carry different conversational functions (acknowledgment versus query) depending on intonation, which our text-only approach cannot distinguish. While we demonstrate that text-based LLM annotation achieves reasonable accuracy, incorporating multimodal information could further improve annotation quality and theoretical coverage.

\section{Ethics Statement}
This research involves annotated dialogue data from the HCRC MapTask corpus, which consists of recordings of human participants engaged in a collaborative task.
The corpus is publicly available for research purposes and doesn't contain any sensitive personal information.

\section{Code and Data Availability}
We release our full code, annotation prompt templates, and the complete annotated dataset to facilitate reproducibility\footnote{{\url{https://github.com/chnln/grounded-misunderstandings-in-maptask}}}.

\section{Acknowledgements}

We appreciate the helpful comments and suggestions from the anonymous reviewers.
We would also like to thank Amy Isard, Matthew Aylett, and Martín Villalba for their valuable input on the analysis of misunderstanding in MapTask during the early stages of this project.
This work is funded by the Dutch Research Council (NWO) through the AiNed Fellowship Grant NGF.1607.22.002, \textit{Dealing with Meaning Variation in NLP}.

\section{References}\label{sec:reference}
\bibliographystyle{lrec2026-natbib_add_date}
\bibliography{references}

\appendix

\section{Appendices}\label{sec:app}

\subsection{Annotation Prompt Template}\label{sec:app-prompt}

Figure~\ref{fig:prompt} in the main text shows an abbreviated version of the annotation prompt template. Below we provide the full content of each section in the prompt template\footnote{To enforce the model's output to adhere to the required format, we not only include the output format section in the prompt, but also attach a JSON-schema file as output format parameter when calling the OpenAI Response API.}. Each section is wrapped in the XML style in the actual prompt, for example, the background section is enclosed by \texttt{<background>} and \texttt{</background>}.

\begin{tcolorbox}[
  colback=gray!5,
  colframe=black!75,
  fonttitle=\bfseries\small,
  left=2mm,
  right=2mm,
  top=1mm,
  bottom=1mm,
  breakable
]
{\small

\tcbsubtitle{Background}
MapTask dialogues involve a route giver guiding a follower using two similar but non-identical maps, and they cannot see each other's map. The giver has the route; some landmarks differ by name (lexical), existence (present on only one map), or multiplicity (appearing twice on one map). You are an expert annotator for reference expressions (REs). Now for each RE, you should annotate the speaker's intended and the addressee's interpreted landmarks using provided landmark IDs. Note that the speaker and addressee are determined by who produces and who receives the RE, not by the fixed giver/follower roles.

\tcbsubtitle{Task Description}
Annotate every RE listed in the \texttt{target\_ref\_ids} section using landmark IDs from the candidate inventory. Also fill in the boolean attributes: \texttt{is\_quantificational} (from the speaker's perspective, whether the expression is used to check or deny the existence of a landmark rather than to refer to a specific one; the following four other attributes are from the addressee's perspective), \texttt{is\_specified} (whether the addressee provides any response that shows they have noticed the RE rather than skipping or ignoring it), \texttt{is\_accommodated} (whether the RE is successfully received and understood by the addressee, false if the addressee mishears, misunderstands, or asks for repetition), \texttt{is\_grounded} (whether the RE is linked to any concrete landmark on the maps by the addressee), and \texttt{is\_imagined} (optional, whether the grounded landmark exists only on the speaker's map). Base every decision strictly on the provided dialogue context and dialogue-act tags.
Provided information includes: (1) dialogue context: dialogue transcript from the beginning of the dialogue until the end of the current transaction (a transaction is a series of moves from one point to another on the maps); (2) maps: both the giver's and follower's maps with map names at the bottom; (3) dialogue act tags: dialogue context annotated with dialogue acts; (4) landmark candidates: a list of all landmarks on both maps. Pick up IDs from the \texttt{landmark\_candidates} section, and do not invent new IDs.

\tcbsubtitle{Landmark ID Explanation}
Landmark ID syntax: \texttt{<map-id>\_<concept>\#<ordinal>@<side>}. \texttt{<concept>} is the landmark name from the original dataset; \texttt{side} $\in$ \{g, f\} indicates which map the landmark is located on; ordinal (0 or 1) is used only when a concept appears twice on the same map; ordinal numbers are ordered from 0 to 1 by their occurrences from bottom to top on the same map, and the number is the same across maps at matching locations. For simultaneous multi-target references, join their IDs with \texttt{+}.

\tcbsubtitle{Annotation Rule}
Follow the annotation workflow for each RE step by step:

\textbf{Step 0}: Identify the speaker and the addressee from the sentence header; fill in the speaker's interpretation (intended landmark ID) first. Exception: if the speaker is querying or denying a landmark previously introduced by the addressee (e.g., ``where is the running river?'' / ``I don't have the running river''), set the speaker's interpretation to empty (``'').

\textbf{Step 1} (\texttt{is\_quantificational}): Set to true when the RE is used only to check or deny existence (e.g., ``Do you have a parked van?'' / ``I don't have any roman baths''); otherwise false.

\textbf{Step 2} (\texttt{is\_specified}): When not quantificational, determine if the addressee shows any uptake of the RE in the context, rather than ignoring or skipping it without reaction; if so, set \texttt{is\_specified=true}, otherwise false. Exception---implicit reuse: if this expression was previously grounded between the interlocutors with no subsequent repair/denial, the current RE may be treated tacitly grounded; set \texttt{is\_specified=true}, and at step (4) mark it as grounded.

\textbf{Step 3} (\texttt{is\_accommodated}): If specified, set to false when the addressee fails to process the RE, e.g., mishearing, confusion, or requesting repetition; otherwise true.

\textbf{Step 4} (\texttt{is\_grounded}): When accommodated, set to true only if the addressee links the RE to a specific landmark on the maps; otherwise false.

\textbf{Step 5} (\texttt{is\_imagined}): When grounded, set to true if the addressee's interpretation is only on the speaker's map; otherwise false.

\textbf{Step 6}: If grounded, fill in the landmark ID for the addressee's interpretation.

\textbf{Step 7}: Provide a concise ($<$50 words) evidence-based reason summarizing the judgement in the reason field for each RE.

\tcbsubtitle{Output Format}
Return strict JSON only:
\begin{verbatim}
{ 
  "dialogue_id": str,
  "landmark_reference_expressions": [
    { 
      "ref_id_unif": str,
      "is_quantificational": bool,
      "is_specified": bool,
      "interpretations": {
        "is_accommodated": bool,
        "is_grounded": bool,
        "is_imagined": bool,
        "giver": str,
        "follower": str
      },
      "reason": str
    },
    ...
  ]
}
\end{verbatim}

\tcbsubtitle{Dynamic Sections}
The remaining prompt sections are instantiated based on all referenced expressions for each request:

\begin{itemize}
  \item \texttt{<target\_ref\_ids>} lists the RE IDs to annotate in the current request;
  \item \texttt{<context\_dialogue>} provides the dialogue transcript from the start of the conversation through the current transaction;
  \item \texttt{<context\_dialogue\_acts>} provides the same context annotated with dialogue acts;
  \item \texttt{<landmark\_candidates>} lists all landmarks on both maps with their \texttt{umlm} IDs.
\end{itemize}
}
\end{tcolorbox}

\refstepcounter{figure}\label{fig:app-prompt}
\begin{center}
\textbf{Figure~\thefigure}: Full annotation prompt template.
\end{center}

\subsection{Reference Chain Dialogue Excerpt}\label{sec:app-chain}

Table~\ref{tab:chain-example} in the main text traces the understanding state transitions for the landmark \emph{vast meadow} in dialogue \texttt{q7ec2}. Below is the relevant dialogue excerpt providing context for those reference expressions (\texttt{refs 112, 114, 115, 116, 119, 127}). Reference expressions are enclosed in \texttt{<<...>>}; the six reference expressions are \textbf{bolded}.

\begin{tcolorbox}[
  title=Dialogue Excerpt: q7ec2 (utterances 131--161),
  colback=white,
  colframe=black!60,
  fonttitle=\bfseries\small,
  left=2mm,
  right=2mm,
  top=1mm,
  bottom=1mm,
  breakable
]
{\small\ttfamily
[giver utt:131] i've got \textbf{<<a meadow>>}\textsuperscript{ref.112} to the left of <<that>> due south pee porashin\\[1pt]
[follower utt:132] \textbf{<<a vast meadow>>}\textsuperscript{ref.114}\\[1pt]
[giver utt:133] \textbf{<<a vast meadow>>}\textsuperscript{ref.115}\\[1pt]
[follower utt:134] i've got \textbf{<<a vast meadow>>}\textsuperscript{ref.116} up just below <<the blacksmiths>> and <<carpenters>>\\[1pt]
[giver utt:135] well is that huge big and there's nothing at all in that space\\[1pt]
[follower utt:136] uh-huh except\\[1pt]
[giver utt:137] about halfway up the page\\[1pt]
[follower utt:138] \textbf{<<vast meadow>>}\textsuperscript{ref.119} about halfway up the page it's empty on mine\\[1pt]
[giver utt:139] well you want to well anyway follow <<crane bay>> round <<the curve>>\\[1pt]
\textnormal{... \textit{(utterances 140--160: participants navigate along crane bay and attractive cliffs)}} ...\\[1pt]
[giver utt:161] just below i've got \textbf{<<a vast meadow>>}\textsuperscript{ref.127} there\\[1pt]
[follower utt:162] right where will i stop that line across the top
}
\end{tcolorbox}

\refstepcounter{figure}\label{fig:app-chain}
\begin{center}
\textbf{Figure~\thefigure}: Dialogue excerpt from \texttt{q7ec2} (utterances 131--161).
\end{center}

\vspace{\baselineskip}
\noindent\textbf{Context}: The giver's map contains two \emph{vast meadow} instances: \texttt{\#0@g} (lower, unique to giver) and \texttt{\#1@g} (upper, shared with follower). The follower has only the upper instance (\texttt{\#1@f}). At \texttt{ref.112}, the giver refers to the lower meadow. The follower's ``a vast meadow?'' (\texttt{utt:132}) is a clarification query. When the giver confirms (\texttt{ref.115}), the follower identifies the upper shared meadow instead, producing a misunderstanding. At \texttt{ref.116}, the follower's spatial description (``below the blacksmiths and carpenters'') shifts the giver's grounding target to the upper/shared meadow, and both align. After a 28-utterance gap discussing other landmarks, the giver returns to ``a vast meadow'' (\texttt{ref.127}) in a new spatial context, but the follower is not able to ground it --- the state reverts to pending.

\end{document}